\pdfoutput=1 

\documentclass[letterpaper, 10 pt, conference]{ieeeconf}  

\IEEEoverridecommandlockouts                              

\usepackage{times}
\usepackage{epsfig}
\usepackage{graphicx}
\usepackage{amsmath}
\usepackage{amssymb}
\usepackage[normalem]{ulem}

\title{\LARGE \bf
Learning Navigation by Visual Localization and Trajectory Prediction
}

\author{Iulia Paraicu$^{1}$ and Marius Leordeanu$^{1,2}$
\thanks{$^{1}$Institute of Mathematics of the Romanian Academy, Calea Grivitei 21, Bucharest, Romania}%
\thanks{$^{2}$Faculty of Automatic Control and Computers, University Politehnica of Bucharest, Splaiul Independentei 313, Bucharest, Romania}%
\thanks{{\tt\small iulia.paraicu@gmail.com},
        {\tt\small leordeanu@gmail.com}}%
}

\begin{document}

\maketitle
\thispagestyle{empty}
\pagestyle{empty}

\begin{abstract}
When driving, people make decisions based on current traffic as well as their desired route. They have a mental map of known routes and are often able to navigate without needing directions. 
Current self-driving models improve their performances when using additional GPS information. Here we aim to push forward self-driving research and perform route planning even in the absence of GPS. Our system learns to predict in real-time vehicle's current location and future trajectory, as a function of time, on a known map, given only the raw video stream and the intended destination. The GPS signal is available only at training time, with training data annotation being fully automatic. 
Different from other published models, we predict the vehicle's trajectory for up to seven seconds ahead, from which complete steering, speed and acceleration information can be derived for the entire time span. Trajectories capture navigational information on multiple levels, from instant steering commands that depend on present traffic and obstacles ahead, to longer-term navigation decisions, towards a specific destination. 
We collect our dataset with a regular car and a smartphone that records video and GPS streams. The GPS data is used to derive ground-truth supervision labels and create an analytical representation of the traversed map. In tests, our system outperforms published methods on visual localization and steering and gives accurate navigation assistance between any two known locations.
\end{abstract}

\section{Introduction}

Not so long ago the idea of self-driving vehicles was more science fiction than a possibility in the near future. Nowadays, advances in deep learning and hardware brought autonomous vehicles closer to reality. The real-time video stream of the surroundings provides the self-driving system with almost the same input that a human would have when controlling a vehicle. However, learning to drive only from video remains a challenging problem as it involves deriving knowledge about a complex world, from a vast quantity of data.
In the literature, there are models \cite{Authors1}\cite{Authors2} that use visual information to extract high-level semantics of the traffic scene and decide the steering action conditioned on these representations. Other works \cite{Authors14}\cite{Authors3}\cite{Authors4} are based on end-to-end models that take as input frames from the video and directly output steering commands. The first approach is easier to interpret by humans, being especially useful to identify and justify failure cases. On another hand, collecting data and training is more efficient in end-to-end solutions, which can learn more relevant features. 

We propose a system that combines end-to-end learning with precise mathematical modeling for automatic visual localization and navigation, thus providing both efficiency and a certain level of explainability. Our model predicts the vehicle's trajectory for the next seven seconds, which provides complete steering and speed information to avoid obstacles and follow a certain route. Our work is thus related to end-to-end learning approaches, which started with the pioneering model Alvinn \cite{Authors12} and continued with the highly successful models in the deep learning era~\cite{Authors14, Authors3, Authors4, Authors9}.

\begin{figure}[t]
\begin{center}
    \includegraphics[width=1\linewidth]{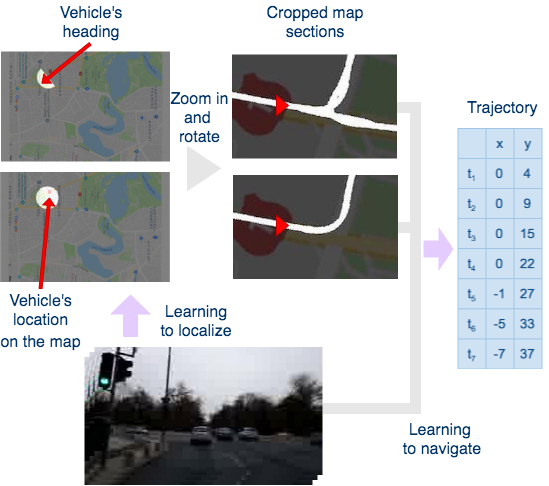}
\end{center}
   \caption{The high-level structure of the system. The first conv net learns to predict the location  on the map by segmentation. Road map segments are cropped around the location, one showing all directions, and the other one only the intended route. From road map crops together with the input frames, the second network predicts the navigation trajectory. The images' map background and the red markers have only an informative purpose.}
\label{fig:system}
\vspace{-0.3cm}
\end{figure}

\paragraph{Visual-based localization} One of the main tasks we tackle is that of visual-based localization, which aims to estimate location from visual input (e.g. images or video). Traditional methods performing this task were based on explicit feature extraction and matching \cite{Authors7}\cite{Authors16}. As with the vast majority of vision problems, the accuracy in localization raised considerably with the advances in deep learning ~\cite{Authors15,Authors24,Authors8, Authors22, Authors23}. Feature-based methods are highly accurate, but often require complex pipelines and are not very robust to changes in weather, lighting conditions and occlusions. Kendall et al.\cite{Authors23,Authors29} introduced the idea of predicting the observer's pose directly from the image with an end-to-end regression CNN. They also use transfer learning from classification nets trained on large datasets such as Places and ImageNet. In their remarkable work, the model achieves good accuracy in real-time, being robust to factors such as unusual lightening or image blur. Marcu et al.\cite{Authors8} proposed an original deep learning approach to localization from images that formulates localization as a segmentation task: the input to a segmentation-style U-Net\cite{Authors19} is a given image and the output is a circle on a map, with the center at the predicted location of that image w.r.t. the map. We compare these methods on our dataset, then propose an improved localization by segmentation solution that achieves strongly superior results to the regression paradigm.

\paragraph{Visual navigation with location information and trajectory prediction}  
The large scale usage of navigation applications such as Google Maps or Waze makes possible to provide both directions and traffic dynamics. Some driving assistance solutions \cite{Authors5}\cite{Authors6} even generate videos from Google Maps images to provide better route recognition to users. 
In the context of autonomous driving Hecker et al.\cite{Authors9} introduced the idea of training end-to-end models with additional online information about the navigation route. Their deep recurrent CNN model receives images from multiple cameras placed at different angles along with a screenshot of a route planning commercial application. 
Amini et al. \cite{Authors10} propose a variational end-to-end solution for navigation and localization that predicts the probability distribution of the vehicle's next pose given an offline map representation, the previous pose, video, and GPS raw streams. 
Another task we tackle is that of trajectory prediction. The idea was first introduced by Glassner et al.~\cite{Authors11}, who developed a trajectory learning model that exceeds a baseline end-to-end steering solution in a simulated highway environment. Their neural network is followed by an analytical module designed to validate the predicted trajectories. Different from the model in~\cite{Authors11}, ours predicts trajectories for a longer time span and is trained and tested on real-world data. Both utility and complexity of our task are increased by the urban scene having intersections where there are multiple possible paths to take. Apart from comprising rich steering information and being easier to interpret, trajectories can also be fitted in the post-processing step to derive smoother steering actions and improve passengers' comfort (studied in \cite{Authors30}). At the higher level, we aim to solve visual-based real-world navigation, by estimating the trajectory without any GPS knowledge, conditioned on the desired destination relative to a map (automatically created during training).
The previous work that best captures the visual learning navigation problem is proposed by Mirowski et al.\cite{Authors28} with a reinforcement learning approach. They solve the maze navigation task in a context closer to real life, by also integrating it with place recognition, which we exploit as well. Their agents successfully learn to reach destinations by taking discrete movement actions in an environment made of Google Street View images. While our supervised learning approach, from automatically annotated data, is very different in terms of data, learning models, mathematical formulation and specific predictions (our trajectory vs. their discrete set of actions), our work and theirs are related at the high level. 

\paragraph{System and data overview}
Our main goal is to replicate humans' capacity to localize and navigate from vision alone with minimal expense in hardware (e.g. smartphone) and training resources.
An overview of the system is presented in Fig. \ref{fig:system}. In the first stage, a deep net predicts the current location from the current frame. In the second stage, the route around the obtained location is fed together with multi-frame visual information to another deep net, which learns end-to-end to predict the future trajectory. Due to the lack of data freely available that is suitable for our proposed approach, we collected our own dataset covering a relatively large area in a European city.
We plan to make all our data and models freely available. We use a regular car and a mobile phone which collects video and GPS streams simultaneously. The ground truth location and steering labels for the video frames are obtained by automatically filtering and processing the GPS stream, thus obtaining annotations at a minimal cost. A system such as the one we propose could be easily deployed at the large scale within a city, for both data collection, annotation, training, and driving assistance, as it requires no specific hardware or manual annotations.

Our \textbf{main contributions} are the following:
\begin{enumerate}
  \item We propose, to our best knowledge, the first deep-learning-based system that simultaneously learns to self-locate and to navigate towards a planned destination from vision only, by exploiting the previous experience of human driving on the same map.
  \item The system is highly scalable at minimal costs and can be easily deployed to learn over an entire city by having it used by many drivers simultaneously. We also introduce the Urban Eastern European Driving Dataset (UEEDD), which we will make publicly available.
  \item Other contributions include: 1) we extend and improve a previous visual localization model and adapt it to learn to localize accurately in traffic. 2) we output trajectories, functions of space vs. time, which comprise steering and speed information for up to seven seconds in the future. 3) the map is created analytically and automatically from the collected GPS data. 
  \item We present competitive numerical results and comparisons to strong baselines and the state of the art.
\end{enumerate}

\section{Creating the Dataset and the Map}

\begin{figure}[t]
\begin{center}
    \includegraphics[width=1\linewidth]{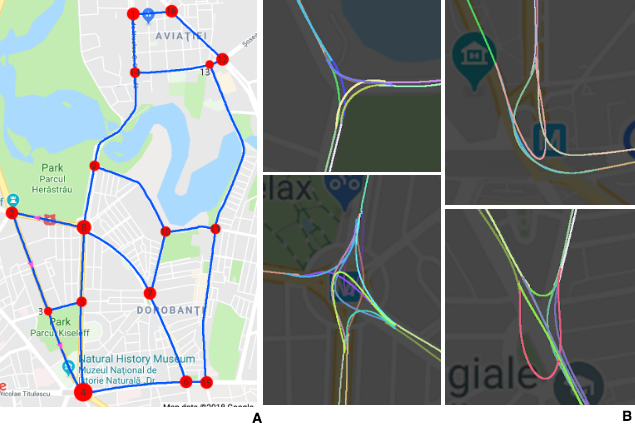}
\end{center}
   \caption{A - The graph structure of the selected driving map. The numbered nodes represent intersections and the edges are the roads connecting them. B - Cropped sections of the analytically obtained map overlapped with the corresponding map regions from Google Maps.}
\label{fig:map}
\vspace{-0.3cm}
\end{figure}

We used a mobile phone to film through the windscreen of the car while driving. A GPS stream was collected simultaneously with the video, by using our own developed Android application.
The map (Fig. \ref{fig:map} A) of the roads covered by the dataset is represented as a directed graph where nodes and edges represent intersections and roads between intersections, respectively. The resulting graph contains 16 nodes and 41 edges, with a total length of 35 km. 
We chose the dataset routes by a simulation designed to cover the graph in a manner that is both uniform and realistic. From the current node, it generates the shortest route to a random destination, then the process is repeated with the previous destination node as the source until a limit of 350 km is reached. The simulation was run 1000 times and the best candidate was selected to be the one with the minimum number of pair edges (two oriented edges connected by a node) that are crossed less than three times. Pair edge crosses are maximized because they depend on the actual route, as an intersection can be passed through in different ways depending on the destination. The UEEDD consists of 19h and 15m of driving videos at 30 fps, taken at different times of the day and traffic conditions. We train our models on 89.72\% of the dataset and test it on the remaining videos of approx. 2h total length. There are 12 continuous video sequences in the test set, distributed uniformly on the map, each of about 10m.

\subsection{Polynomial Based Analytical Techniques}
\label{sec:poly_fit}
Due to the large data space and noise in the data, conv nets sometimes express undesired and unpredictable behavior. Precise mathematical models can often complement the NN ones, to reduce such problems and improve generalization. In this paper, we will employ an analytical approach based on polynomial functions of time or distance, used for fitting time trajectories (functions of time) and map segments (paths - functions of distance), respectively. We chose the mathematical model of polynomials because they are flexible and also simple to compute~\cite{Authors31}. The following method, presented for trajectories was also employed in the case of map segments.

Given the initial 2D trajectory points, each trajectory can be analytically modeled with two polynomial functions of time for the $x$ respectively $y$ space components. We exemplify the 3rd order case, for which we solve for the polynomial coefficients.

\begin{align}
  x(t) &= a_1t^3 + a_2t^2 + a_3t + a_4\\
  y(t) &= b_1t^3 + b_2t^2 + b_3t + b_4.
\end{align}

As the x-y coordinates are known along with their time steps, the coefficients can be estimated using the method of linear least squares, by forming the data matrix $\mathbf{T}$, with values $t_i^3, \& t_i^2 \& t_i$ on the i-th row and the $\mathbf{x}$ and $\mathbf{y}$ vectors containing the target $x(t_i)$ and $y(t_i)$ values. Then the optimal coefficient vectors $\mathbf{a}$ and $\mathbf{b}$ of the convex least squares problem are
found using the classic formula: $\mathbf{a = (T^TT)^{-1} T^Tx}$ and $\mathbf{b = (T^TT)^{-1} T^Ty}$. After obtaining the polynomial coefficients we could analytically find $x(t)$ and $y(t)$ for any time step $t$. The analytical approach ensures that the resulting trajectories are smooth and makes possible to sample XY points in the future at equal time intervals.

\subsection{Analytical Map Representation}

Visual map representations can improve the prediction of a driving neural net, especially in intersections, as shown in~\cite{Authors9,Authors30,Authors10}, who provide their models with visual map captions around the current GPS location. Different from them, we analytically derive a map from the collected GPS streams of the routes traversed in the training phase. Then we provide our trajectory prediction model the analytical map information around the current predicted location.

The key idea in map creation is fitting polynomials on conveniently chosen compact groups of location samples to obtain curved map segments, that together form the entire map. We choose the map segments as follows:

\begin{itemize}
  \item Each directed edge of the graph (connecting road) represents a segment.
  \item Nodes (intersections) include multiple segments. Each node segment is defined by a distinct combination of the two connected pair edges and their direction.
\end{itemize}

The map creation algorithm takes as input lists of geolocation samples for every route, a complete list of destinations (nodes) in order, the geolocation of the nodes centers and their radius. It also implies that every two consecutive map segments along a route continue one into the other smoothly. The steps for building the map representation are:

\begin{enumerate}
  \item Place each geo-coordinate sample into a bucket corresponding to its segment together with the distance from it to the start of the segment. The distance to the segment start is 0 when entering the segment, then for each following sample point, its associated distance is the previous point distance summed with the euclidean distance between the two samples.
  \item Fit polynomial functions of distance for each segment given the points and the corresponding distances in its bucket, using the method presented in Sec. \ref{sec:poly_fit}, but using $x(d)$ and $y(d)$ instead of  $x(t)$ and $y(t)$, where \textit{d} is the known distance to the start of the segment. The degree of polynomials is directly proportional with the length of the modeled segments but significantly smaller than the number of points.
  \item Using the analytical model we can now sample points at 1m distance interval along each segment from $s=0$ to the segment's end. After this step, some pairs of segments (with points sampled in their analytical form) will have small gaps in between and they are not guaranteed to connect smoothly. To tackle this we consider, for each segment the ending parts of its neighboring map segments and refit its polynomial function, to obtain a final smooth and continuous map representation (Fig. \ref{fig:map} B), by sampling from $s=-delta$ to $s=end+delta$ for each segment, where $delta$ is a small distance buffer. 
\end{enumerate}

\section{Learning to Localize}
\label{sec:localization}

Our system has to recognize the current location of the automobile from a continuous online video stream. We test two previously proposed approaches, one formulating localization as regression and the other as segmentation, then extend the last one and obtain significantly better performances.

\subsection{Localization by Regression}

In \cite{Authors23, Authors29} authors train a regression NN to predict the e 6-DOF camera pose from single images of landmarks. The model's output consists of two vectors: the 3-component position vector of distances on the axes \textit{x}, \textit{y} and \textit{z}, and the 4-component quaternion orientation vector that represents the rotation around the three axes. The loss function consists of a weighted sum of the L2 norms between each of the outputted vectors and their corresponding ground-truth. In \cite{Authors23} the weighting is done by hyperparameters, whereas in \cite{Authors29} the weights are learned based on the hypothesis of the task homoscedastic uncertainty \cite{Authors32}.
To express the vehicle's pose on a 2D map we only need a 3-DOF representation. When implementing the models above we keep exactly the same setting and evaluate only the pose components of interest, the other ones being constant. The results of the regression methods are analyzed in Sec. \ref{sec:localization_results}.

\subsection{Localization by Segmentation}

The idea of learning localization by segmentation was introduced in \cite{Authors8} for the case of satellite images with associated geolocation (2-DOF pose). Their two-stage model predicts the mask of a dot on an output map (representing the geographic map), such that the center \textit{(x,y)} of the dot represents the coordinates of the image location w.r.t to the geographic map. Treating geo-localization as a segmentation problem has the advantage that segmentation nets capture well the relation between "what" and "where", between semantics and geometrical relations in the output space. Also in the case of complex output distributions, a segmentation net can output several possible locations (e.g. output several dots, later post-processed for getting a final answer), whereas regression is directly forced to produce a single answer.

The second stage module in \cite{Authors8} is the basis for our architecture. We adapt it to predict 3-DOF poses end-to-end from RGB input images. Details about the network's architecture and training are given in \cite{Authors8}.
We add a second output map to also predict the orientation, on which only half of the dot is segmented towards the vehicle's heading direction. The orientation is obtained from the vector connecting the center of the dot on the first map with the center of the half-dot on the second.  The net is able to localize with high accuracy, except for some isolated sections where large errors are made or the output dot is missing completely for tens of seconds in a row. We looked at the failure cases and discovered that most miss-predicted images contain occlusion elements, such as large vehicles right in front of the car.

\subsection{A Deeper Look into the Segmentation Model}

\begin{figure}[t]
\begin{center}
    \includegraphics[width=0.9\linewidth]{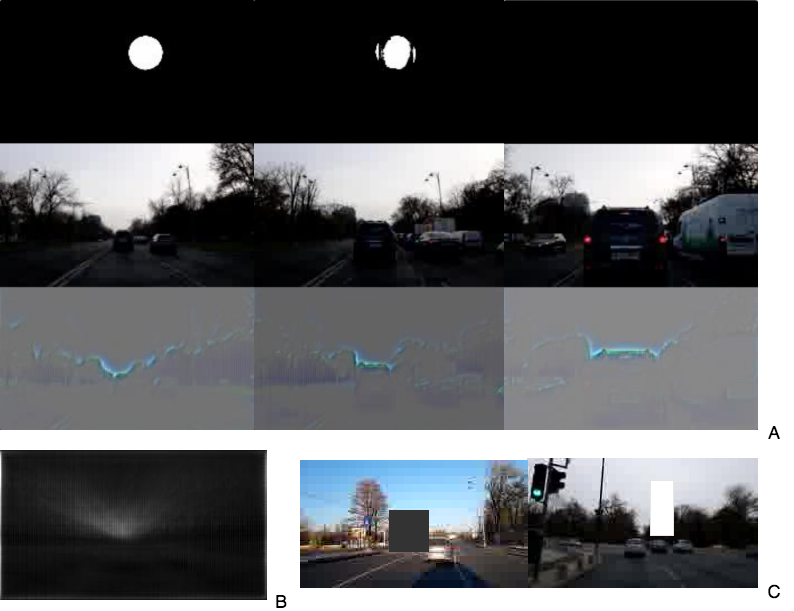}
\end{center}
   \caption{Analyzing the network's attention by applying the guided backpropagation technique. A - The top row shows the prediction of the network for the observer's location, while the bottom one shows the visual cues of most interest for the network. It can be seen that as the camera approaches the SUV vehicle, the skyline shape changes and the segmentation of the dot fades away. B - The guided backpropagation mean activation values over pixels in training set images. C - Images augmented by our algorithm.}
\label{fig:guided_prop}
\vspace{-0.3cm}
\end{figure}

For most segmentation tasks the actual segmented area in the image also corresponds to the region of interest, relevant for producing its segmentation. However, this observation does not apply to localization by segmentation, for which there is no direct logical relation between positions in the input to the position of the dot in the output map. However, there are methods~\cite{Authors25,Authors26} in the literature that discover which locations in the visual input are relevant for a particular network's output response, based on the activation values of the neurons in forward and/or backward passes. The guided backpropagation\cite{Authors25} algorithm fits best our case, as it outputs a fine-grained relevance map of the input image and it works with any CNN model.

The saliency maps obtained by guided backpropagation show clearly that the model's attention is highly focused on the contour of the skyline, with a higher weight on the central region of the image (Fig. \ref{fig:guided_prop} A).  When a taller vehicle obstructs parts of the skyline, the output immediately degrades: the segmented dot loses its shape, it is miss-placed or even disappears. To avoid this phenomenon we must consider visual parts of the image that are less affected by such distractors. 
For this, we propose an informed technique of data augmentation during training. Firstly, the mean of each input pixel's (guided backpropagation) activation values is computed (Fig. \ref{fig:guided_prop} B). From all calculated values over the training set an empirical distribution is obtained. Thus, positions in the image corresponding to pixels with higher probabilities are more likely to influence the network's predictions. For each training image, we randomly sample from the distribution a pixel location in which we center a randomly sized box of constant random gray-scale value (to mimic a large vehicle in front of the camera). The proposed augmentation method significantly improves accuracy.

\subsection{Experiments and results}
\label{sec:localization_results}

We evaluate the response rate of the network (how often it produces a dot), as well as the mean and the median location errors, expressed in meters (see Tab. \ref{tab:localization_evaluation}). For the segmentation approach, we only take into consideration predicted dots with an area between 25\% and 175\% of the area of the perfect dot of 15px radius.

\begin{table}
\begin{center}
\begin{tabular}{|c|c|c|c|c|c|c|}
\hline
 & \multicolumn{3}{|c|}{Position Error (m)} & \multicolumn{3}{|c|}{Orientation Error (deg)} \\
\hline
Method & Response & Mean & Median & Response & Mean & Median \\
\hline
\cite{Authors23} & \textbf{100\%} & 58.09 & 17.75 & \textbf{100\%} & 7.81 & 2.41 \\
\cite{Authors29} & \textbf{100\%} & 50.84 & 15.39 & \textbf{100\%} & 7.33 & 1.88 \\
\cite{Authors8} & 91.00\% & 27.36 & 11.44 & - & - & - \\
Ours1 & 94.72\% & 17.31 & 11.55 & - & - & - \\
Ours2 & 96.35\% & 16.89 & 11.18 & 96.08\% & \textbf{3.65} & 1.43\\
\hline
Ours3 & \textbf{100\%} & \textbf{16.05} & \textbf{10.90} & \textbf{100\%} & 3.73 & \textbf{0.67} \\
\hline
\end{tabular}
\end{center}
\caption{Results for 3-DOF pose prediction}
\label{tab:localization_evaluation}
\vspace{-0.7cm}
\end{table}

It can be seen that segmentation approaches \textit{\cite{Authors8}}, \textit{Ours1}, \textit{Ours2}, \textit{Ours3} perform clearly better than the regression ones \textit{\cite{Authors23,Authors29}} in terms of pose precision, evincing that segmentation formulation is superior. Also, the PoseNet method proposed in 2017~\cite{Authors29} outperforms the earlier one from 2015~\cite{Authors23}, as expected.
The proposed data augmentation, \textit{Ours1}, raises the response percentage of \textit{\cite{Authors8}} setup with 3.72\% and decreases the mean error with about 10m (the number of outlier responses is greatly reduced), while the median error remains the same. When the network is modified to perform both position and orientation prediction (\textit{Ours2}), the results for the position task alone rise, so the response rate is 1.33 higher and the errors are slightly lower. The performance also improves when we project the predicted locations on the analytical map segments (\textit{Ours3}). We first project a location onto every segment, then replace it with its projection on the closest segment. By applying this step we also obtain associated poses for all examples in the test set. After projection, we perform additional time smoothing of the position by locally fitting time polynomials.

\section{Learning to Navigate}

The final stage of our method uses the analytical map and the visual localization net, to learn how to navigate in dynamic traffic between previously seen places. We predict trajectories to train a model that is better capable of following a route. It produces significantly fewer high-frequency oscillations and local errors than models with simple steering output.

\begin{figure}[t]
\begin{center}
    \includegraphics[width=1\linewidth]{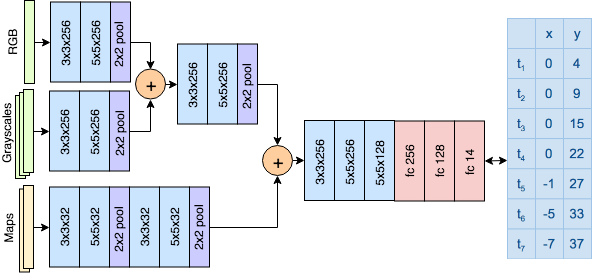}
\end{center}
   \caption{The architecture of the navigation net.}
\label{fig:guided_prop}
\vspace{-0.3cm}
\end{figure}

The navigation net architecture (Fig. \ref{fig:navigation_net}) is an adaptation of the model described in~\cite{Authors20}. The inputs to the network come along three branches: one is for the current RGB frame, another branch takes input three grayscale frames sampled uniformly from the last two seconds and the third has input two binary images of the analytical road map. The first map contains all roads around the current location, while the second road map only shows the road segments that are part of the vehicle's route towards the desired destination. The width of a pixel in the map images represents 1m in the real world. The map is limited to a local neighborhood centered at the predicted vehicle location.
The model outputs 7 pairs of real (x,y) values corresponding to points coordinates defining the trajectory, one per second. The training loss function  is the mean of the euclidean distances between predicted and truth points ($N=7$):

\begin{align}
  L &= \dfrac{\sum_{n=1}^{N}\sqrt{(x_n^t - x_n^p)^2 + (y_n^t - y_n^p)^2}}{N}
  \label{eq:loss}
\end{align}

\subsection{Experiments and Results}

From the prediction of trajectories, we can evaluate the network's performances at both steering and following route directions. Since long-time movement prediction is an unexplored subject in current autonomous driving research, we cannot evaluate our method directly against other solutions. In the following experiments, we modify a baseline~\cite{Authors3} end-to-end steering model (by Nvidia) and the state-of-the-art one~\cite{Authors9} to output steering commands for 7 time steps instead of just one. As the model in \cite{Authors9} has a separate branch for visual navigational information, we provide it the same two concatenated map images, like in the case of our model. The baseline model form Nvidia however, learns only from single frames, as it was designed for. 


\begin{table}
\begin{center}
\begin{tabular}{|c|c|c|c|c|c|c|c|c|}
\hline
\multicolumn{8}{|c|} {MAE Speed (m/s)} \\
\hline
 & 1s & 2s & 3s & 4s & 5s & 6s & 7s\\
\hline
\cite{Authors3} & 1.9 & 1.91 & 1.99 & 1.94 & 1.96 & 1.95 & 2.33 \\
\cite{Authors9} & 1.76 & 1.7 & 1.68 & 1.68 & 1.69 & 1.72 & 1.76 \\
Ours & \textbf{0.94} & \textbf{0.92} & \textbf{0.91} & \textbf{0.92} & \textbf{0.94} & \textbf{0.98} & \textbf{1.03} \\
\hline
\multicolumn{8}{|c|} {MAE Steering Angle (deg)} \\
\hline
 & 1s & 2s & 3s & 4s & 5s & 6s & 7s\\
\hline
\cite{Authors3} & 1.01 & 1.61 & 2.09 & 2.65 & 3.14 & 3.9 & 5.48 \\
\cite{Authors9} & 0.91 & 1.32 & 1.74 & 2.05 & 2.39 & 2.95 & 4.3 \\
Ours & \textbf{0.84} & \textbf{1.26} & \textbf{1.68} & \textbf{2} & \textbf{2.36} & \textbf{2.91} & \textbf{4.24} \\
\hline
\end{tabular}
\end{center}
\caption{Mean errors for speed and steering angle (lower is better)}
\label{tab:steering_error}
\vspace{-0.7cm}
\end{table}

\begin{figure}[t]
\begin{center}
    \includegraphics[width=1\linewidth]{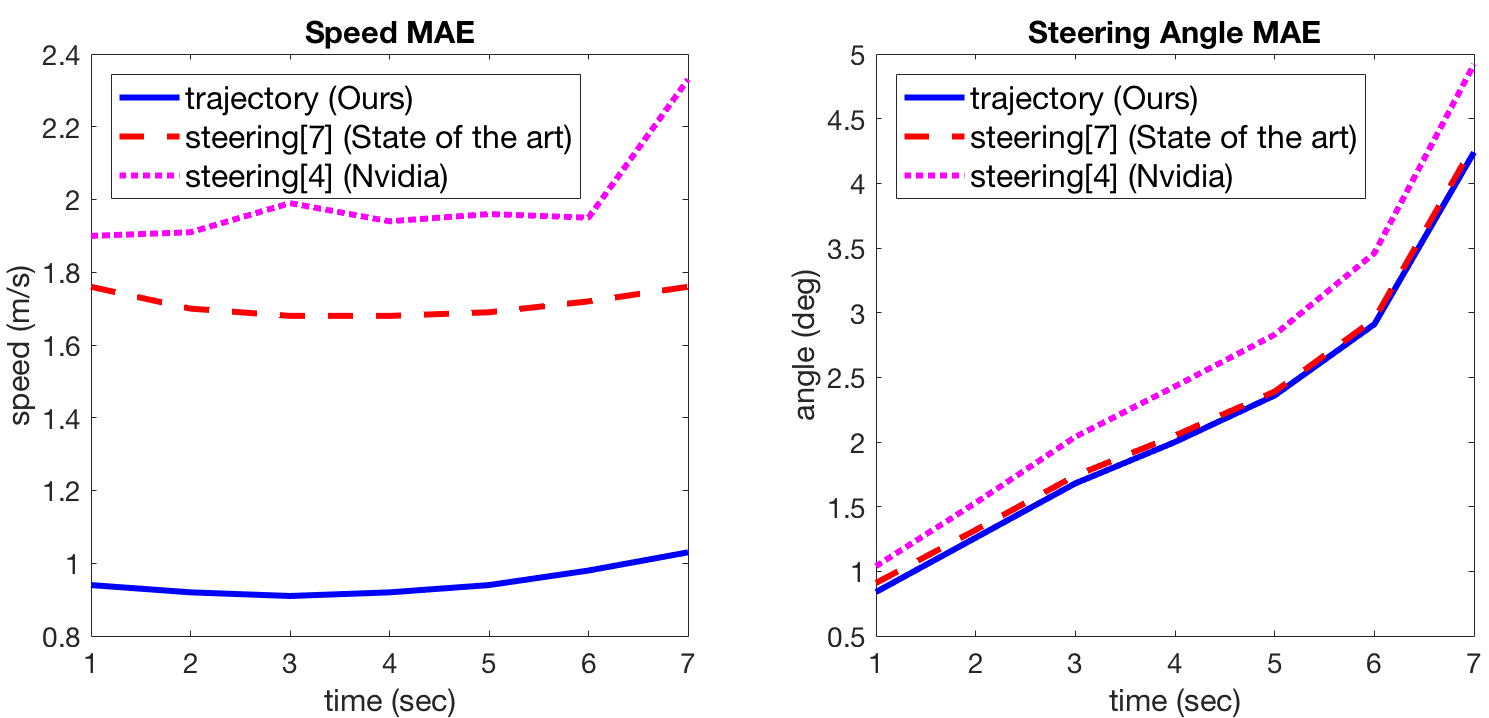}
\end{center}
  \caption{MAE for speed and steering angle. All methods mostly maintain the prediction quality of speed as the time interval expands, but our network has a significantly lower magnitude of errors. The angle errors of our method and the one in \cite{Authors9} are very close, whereas the ones for \cite{Authors3} are higher.}
\label{fig:steering_error}
\vspace{-0.3cm}
\end{figure}

First, we evaluate the steering performance in the same manner as it is done in the literature, but over multiple time steps in the future. The average control commands of speed and angle on seven time intervals of one-second length are derived from the trajectory output of our navigation net. In Tab. \ref{tab:steering_error} we present experimental comparisons with the steering methods w.r.t. steering angle and speed performance. For the speed prediction, our model achieves significantly lower errors, than both steering ones. The steering angle errors of our method and the one in \cite{Authors9} evolve tightly together over time, whereas the ones for \cite{Authors3} are higher.


\begin{figure}[t]
\begin{center}
    \includegraphics[width=0.75\linewidth]{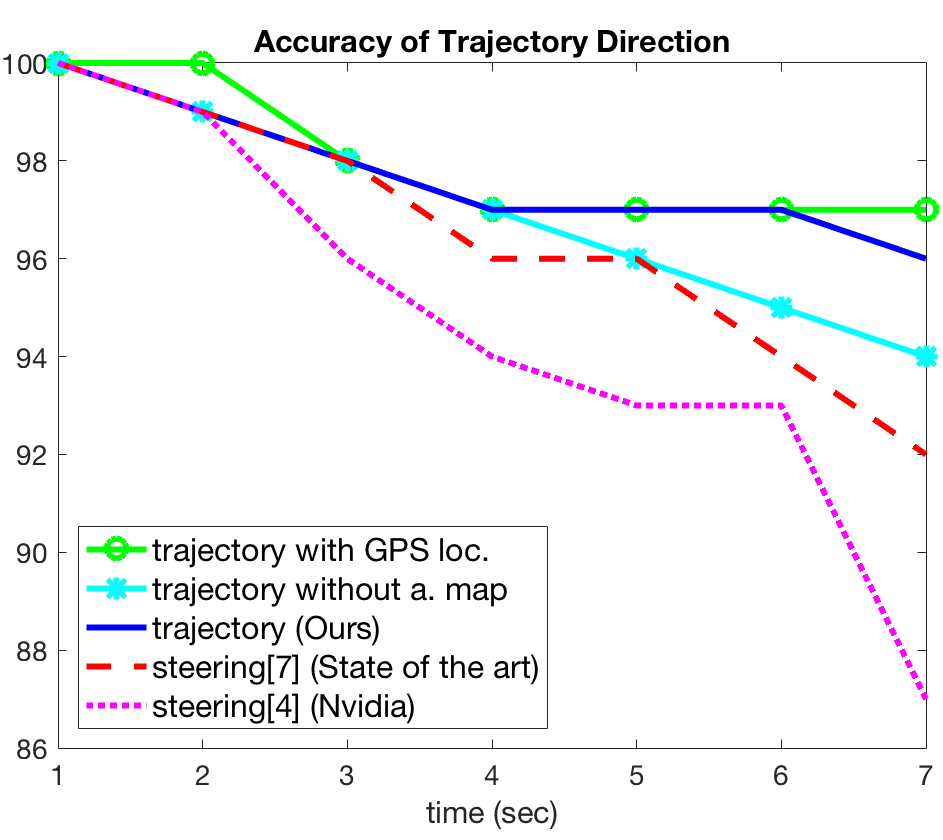}
\end{center}
  \caption{The accuracy of direction in intersections for trajectory points (higher is better). The percentage of trajectory points predicted on the correct route, is mostly higher for our models, than for the steering ones. Note that the accuracy of borderline points is harder to establish because of not having precise location and road structure information available. However, the trajectory points that are clearly off the road (present at broader time intervals due to accumulated errors) are easily spotted.}
\label{fig:trajectory_dir_acc}
\vspace{-0.3cm}
\end{figure}

\begin{figure}[t]
\begin{center}
    \includegraphics[width=1\linewidth]{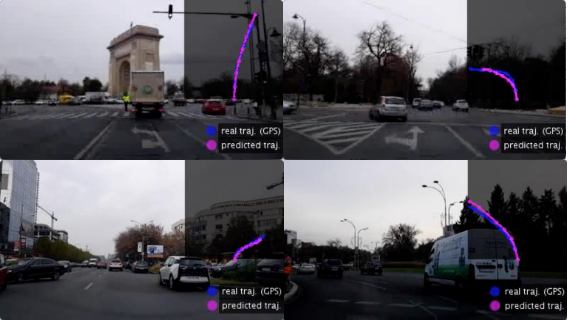}
\end{center}
   \caption{The trajectory predicted by the model in various traffic situations: before entering a roundabout, just before turning left, beginning to turn right and keeping left steer in a roundabout.}
\label{fig:traj-demo}
\vspace{-0.3cm}
\end{figure}

To better evaluate the navigation capabilities of the competing models we introduce a direction performance metric, for when the vehicle is an intersection and different paths may be correct depending on the final destination. At a given moment in time, each approach (either based on trajectory prediction or direct multi-step steering and speed prediction) estimates the location of the vehicle for each of the next time steps. We are interested in predicting locations that are closer to the road segment that belongs to the correct route. By projecting the predicted locations on the road map segments, we can immediately estimate how often the correct route is the closest to the predicted locations among several possible routes.
In Fig. \ref{fig:trajectory_dir_acc} we show the directions accuracy rates (the percentage of times the directions are correct when in an intersection), at each second, for our approach vs the steering ones. We notice a drop in performance as the time span increases, with our net having the lowest rate of decrease compared to \cite{Authors9} and \cite{Authors3} (which has by far the worst performances). We also compare setups similar to \textit{Ours}, to better understand the impact of using the analytical map and the localization prediction (vs. true GPS signal) in the navigation performance. As expected the accuracy drops a bit when navigation is learned without using the analytical map at the input. Also, the performance improves when the true GPS signal is used. However, in all these cases the performances of the models based on our trajectory prediction approach are similar or better than the ones of the steering models. The results demonstrate the value and robustness of learning to predict trajectories over an extended time, our model being state of the art for the end-to-end vehicle steering task. In Fig. \ref{fig:traj-demo} we present a few qualitative trajectory results during test time, which illustrate the ability of our model to handle relatively complex situations and also agree with the ground truth.

\section{Conclusions}

We propose a low-cost and self-contained system capable to learn to localize and navigate using visual input alone.
To our best knowledge, we are the first to propose such a system that requires only a smartphone during the automatic data acquisition, annotation, and testing. We also introduce a mathematical method for constructing an analytical road map from the collected data and introduce a large dataset collected and mapped in this manner.
We further extend a strong approach for localization by segmentation and also develop an intelligent data augmentation procedure that generates training samples for robust learning in the presence of large vehicles on the road. Overall, we explore end-to-end autonomous driving in a top-down fashion, by learning to predict navigation trajectories conditioned on the final destination, from which we derive steering angle and speed values. Our results are state of the art, for prediction of low-level steering control as well as at the higher level of driving directions. We put it all together into a complete inexpensive package that can perform all necessary steps, with minimal human intervention for learning to localize and navigate. Our approach is scalable and competitive and for these reasons, it has the potential to become a strong alternative to current solutions for navigation assistance.


\textbf{Acknowledgements:} This work was supported by UEFISCDI, under Projects  EEA-RO-2018-0496 and PN-III-P1-1.2-PCCDI-2017-0734.


{\small
\bibliographystyle{ieeetr}
\bibliography{root}
}

\end{document}